\documentclass[letterpaper]{article} 
\usepackage{times}  
\usepackage{helvet}  
\usepackage{courier}  
\usepackage[hyphens]{url}  
\usepackage{graphicx} 
\usepackage{caption} 
\usepackage{algorithm}
\usepackage{algorithmic}
\usepackage{multirow}

\usepackage{todonotes}

\usepackage{amsmath}

\usepackage{newfloat}
\usepackage{listings}
\DeclareCaptionStyle{ruled}{labelfont=normalfont,labelsep=colon,strut=off} 
\floatstyle{ruled}
\newfloat{listing}{tb}{lst}{}
\floatname{listing}{Listing}
\title{Learning with Local Gradients at the Edge}
\author{
    Michael Lomnitz\textsuperscript{},
    Zachary Daniels\textsuperscript{},
    David Zhang \textsuperscript{},
    Michael Piacentino \textsuperscript{}
}

\begin{document}

\maketitle

\begin{abstract}

To enable learning on edge devices with fast convergence and low memory, we present a novel backpropagation-free optimization algorithm dubbed Target Projection Stochastic Gradient Descent (tpSGD). tpSGD generalizes direct random target projection to work with arbitrary loss functions and extends target projection for training recurrent neural networks (RNNs) in addition to feedforward networks. tpSGD uses layer-wise stochastic gradient descent (SGD) and local targets generated via random projections of the labels to train the network layer-by-layer with only forward passes. tpSGD doesn’t require retaining gradients during optimization, greatly reducing memory allocation compared to SGD backpropagation (BP) methods that require multiple instances of the entire neural network weights, input/output, and intermediate results.
Our method performs comparably to BP gradient-descent within 5\% accuracy on relatively shallow networks of fully connected layers, convolutional layers, and recurrent layers. tpSGD also outperforms other state-of-the-art gradient-free algorithms in shallow models consisting of multi-layer perceptrons, convolutional neural networks (CNNs), and RNNs with competitive accuracy and less memory and time.
We evaluate the performance of tpSGD in training deep neural networks (e.g. VGG) and extend the approach to multi-layer RNNs. These experiments highlight new research directions related to optimized layer-based adaptor training for domain-shift using tpSGD at the edge.

\end{abstract}

\section{Introduction}

AI-based systems operating in rapidly changing environments need to adapt in real-time. The adaptability of the current edge devices is limited by Size, Weight and Power (SWaP) constraints, the memory wall from current system architectures \cite{McKee2011}, and the lack of efficient on-device learning algorithms. As a result, current AI systems adapt by retraining on the cloud due to the overwhelming complexity of AI models. Adaptation, or domain transfer for a new data distribution, often requires fine-tuning of the entire or partial network (e.g., last few layers). Similarly, it is often necessary to perform conversion from full precision \cite{raghavan2017bitnet} to fewer bits in order to fit the limited hardware architecture, and doing so might require updating the network by distillation. Despite the fact that gradient-descent (GD) backpropagation (BP) learning is one of the most studied and successful approaches, the dominant approach still exhibits some limitations, including vanishing and exploding gradients \cite{difficultyoftraining}, difficulty with data quantization, and inability to handle non-differentiable parameters \cite{luizerothorder}. Extending GD BP to on-device training exhibits several additional problems related to limited memory availability: 1) The weight transport problem requires each layer have full knowledge of other weights in the neural network (NN); 2) The update locking problem requires a full forward pass before the feedback pass, leading to the high memory overhead. In GD BP, all layer inputs and activations in both forward and backward passes need to be buffered before weight updates are complete. 

\begin{figure}
  \centering
  \includegraphics[width=0.8\columnwidth]{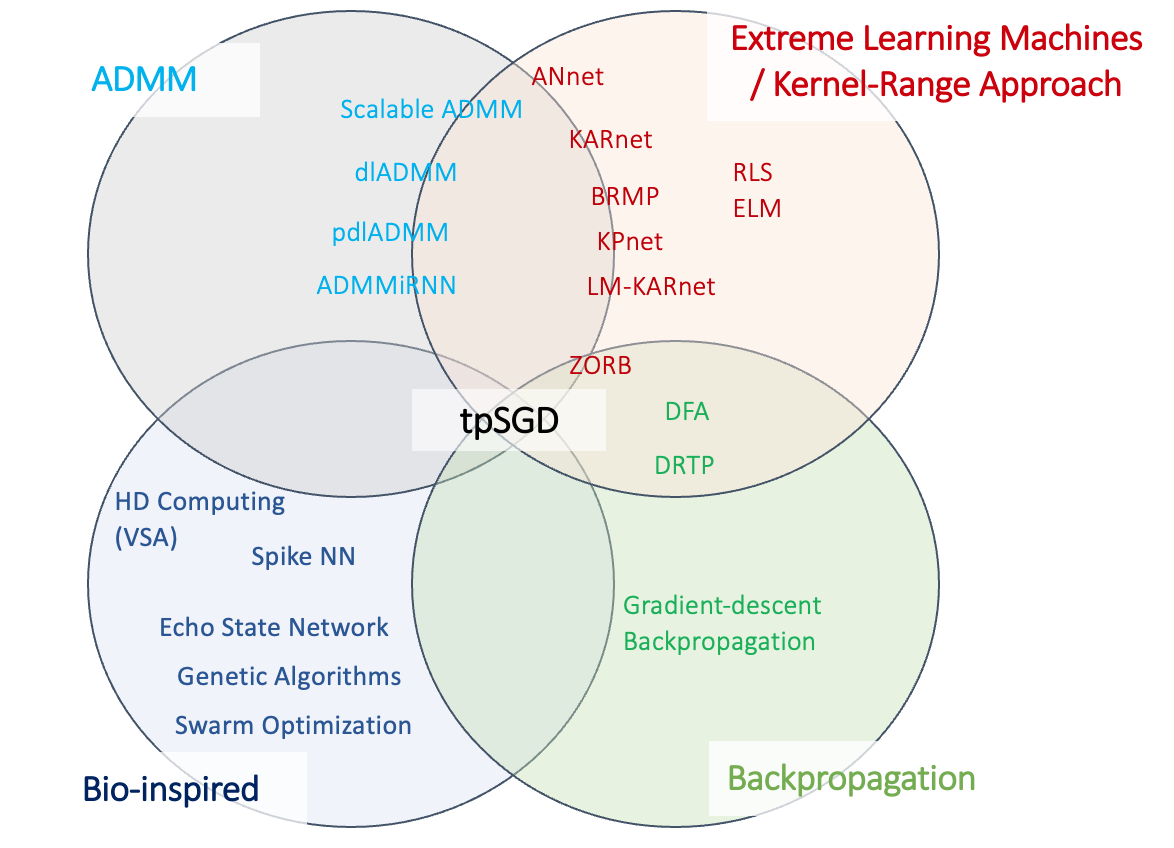}
  \caption{Diagram showing different families of learning algorithms.}\label{fig:conv_projection}
\end{figure}

To overcome issues associated with GD BP, especially for training at the edge, many gradient-free (GF) methods, as summarized in Figure 1 (Bio-inspired, ADMM, and ELM/KR), have been studied. Early GF approaches such as Genetic Algorithms, Swarm Optimization, ADMM, etc. solved the learning problem without using GD. However, they do not show distinctive energy-time saving nor are they memory-efficient and scalable. Biologically-plausible approaches such as Hyperdimensional (HD) computing, and Spike NN (SNN) exhibit energy and/or memory-efficient learning; nevertheless, training these binary represented weights and inputs is most successful based on the approximation from the floating point NNs with distillation. Alternating Direction Method of Multipliers (ADMM) \cite{admm} decomposes the network training into a sequence of sub-steps that can be solved as simple linear least-squares problems, e.g., using Moore-Penrose (MP) pseudoinverse. ADMM optimizes each layer using the following layer as a condition (Lagrange multiplier). As such, it is a forward pass-only algorithm. ADMM variants (Scalable ADMM, dlADMM, pdADMM \cite{taylor2016training, wang2019admm, wang2020pADMM}) have tried to efficiently utilize available hardware resources and reduce compute time. However, they take many iterations to converge to the final solution. Although relatively mature and scalable to large problems, the learning time of ADMM is still much longer compared to other GF methods such as KPnet \cite{zhuang2020training}.

Most of the gradient-free approaches have demonstrated their performance in shallow NN models and in fully connected layers. They are less accurate than deep NN models trained with BP nor are they scalable. However, the combination of these approaches shows a promising trend. For example, KPnet combines Extreme Learning Machines (ELM) \cite{huang2011extreme} with label encoding and target propagation to speed up training and improve scalability. LM-KARnet\cite{zhuang2020training} trains with time savings of 10-100X in shallow networks. ZORB \cite{ranganathan2020zorb} is another GF training algorithm that combines rapid training of ELMs and MP pseudoinverse. It is 300X faster than the Adam optimizer on shallow networks and also achieves comparable accuracy.

Frenkel et al. proposed Direct Random Target Projection (DRTP) \cite{Frenkel_2021, nokland2016direct}, a method introducing target projection into GD learning algorithms without feedback. DRTP alleviates two key BP issues (weight transport problem and update lock) by enabling each layer to be updated with local information as the forward evaluation proceeds. It allows training hidden layers at low computational and memory costs. The locality concept was proven to work on 1-2 shallow layers, and the implemented DRTP in an event-based CNN processor \cite{frenkel2020snn} required only 16.8\% power and 11.8\% silicon area overheads. Based on this concept, we propose a novel backpropagation-free optimization algorithm called \emph{target projection Stochastic Gradient Descent (tpSGD)}, capable of training NN models at the edge. tpSGD trains neural networks layer-by-layer using only forward passes. As far as we know, tpSGD is the first method to extend local training to arbitrary loss function, use target projection for training RNNs, and scale up the locality concept to deep neural networks (DNNs). More specifically, we do not focus on finetuning the entire network at the edge.  Instead we study whether tpSGD can finetune optimized shallow layers to adapt to new environments with less memory usage, more efficient computing, and faster convergence. In summary, tpSGD has the following unique features:

\begin{itemize}
    \item Local efficient computation per layer using SGD and target projection. Our method performs comparably to GD BP (within 1\% accuracy) on relatively shallow networks with 1-6 trainable convolutional and/or fully connected layers, while eliminating the need to retain and compute through the network in the backward pass.
    \item Extends target projection to multichannel convolutional layers via filter-based sampling, enabling CNN training.
    \item Extends target projection to recurrent neural networks (RNN) (standard, stacked multi-layer RNNs, and bidirectional RNNs), demonstrating non-trivial learning and in many cases, performing within 5\% accuracy of BP-trained networks on several benchmark datasets.
    \item Demonstrates state-of-the-art (SoA) performance and decreased training time compared to other BP-free algorithms (e.g., dlADMM, ZORB, KARnet).  
    \item Scalable to DNNs by training the optimized shallow network adaptors that connect to the fixed DNN encoders. 
\end{itemize}

\section{Methodology} 
  
\subsection{Target Projection and Target Propagation}
In target propagation \cite{ranganathan2020zorb}, the labels are ``propagated'' backward through the network by sequentially inverting each operation (layers and activations) in order to obtain the corresponding output to the labels. For instance, for a linear layer $z_i = x_i W_i$, this involves obtaining an approximate value for the inverse of the layer weights $W_i^\dagger$ via the MP pseudoinverse given $x_i$ as input from the output of the previous layer $i-1$ and $z_i$ as the output features of layer $i$ target propagated from the labels through inversion of the final $N-i$ layers. 
  
In target projection \cite{kpnet,Frenkel_2021}, one-hot encodings of the labels are directly projected to a given layer during the optimization step.  Given an intermediate layer $L_{i}$ in a NN, we generate local targets $y_i$ for the layer by projecting the data labels $y^{*}$ via a random projection matrix ($P_i$). The focus of tpSGD is to provide a fast and scalable BP-free algorithm for training NNs, so we chose target projection, replacing the need to invert all of the layers and activations after a given layer, with a single matrix multiplication.

\subsection{Target Projection SGD}
   tpSGD is designed with the goal of scaling BP-free training to larger datasets and deeper networks required for complex tasks.  Unlike ZORB \cite{ranganathan2020zorb} or LM-KARnet \cite{zhuang2020training}, we first utilize GD-based optimization for the individual layers instead of MP pseudo-inverse which optimizes weights on an entire data set at once. While ZORB and LM-KARnet struggle to process the entire CIFAR dataset (especially when convolutions are included) our algorithm can process them in batches providing more freedom to extend to datasets and tasks with larger memory requirements. Second, and in contrast to BP, which  calculates a full forward and backward pass through the network, tpSGD is a feedforward only optimization algorithm. In tpSGD, we train each layer in the network sequentially starting from the layers closest to the input.  For a given layer $L_i$ in a NN with $i\in N$ layers, the input to that layer $x_i$ is obtained by running the forward pass over all previous $j = 1$ to $i-1$ layers.  The target output $y_i$ is obtained via a random projection of the one-hot encoding of the data labels.  The input $x_i$ and projected targets $y_i$ are used to train the layer  using Adam optimizer and the Mean Squared Error (MSE) between the predictions and $y_i$ (either before or after activation). Once a layer is trained, we fix the weights and move on to the next layer following the same approach until the final layer is reached as we no longer require a projection.
  The next sections will discuss connections to existing projection-based training algorithms before moving on to the the details of generating the random projection matrices $P_i$ for the different supported trainable layers: dense, 2d convolutions and recurrent.

\subsection{Connections to Existing Projection-Based Training Algorithms}

\begin{table*}
  \centering
  \begin{tabular}{|c|c|l|}
    \hline
     Method & $J_{i}(.)$ & \multicolumn{1}{|c|}{$w_{i}^{(t+1)}$} \\ \hline
     DFA & Cross-Entropy($y^{*}$, $y_{N}$) & $w_{i}^{(t)} + \eta P_{i}(y^{*} - y_{N})\odot\sigma'_{i}(z_{i})$  \\
     sDFA & Cross-Entropy($y^{*}$, $y_{N}$) & $w_{i}^{(t)} + \eta P_{i}\text{sign}(y^{*} - y_{N})\odot\sigma'_{i}(z_{i})$ \\
     DRTP & Cross-Entropy($y^{*}$, $y_{N}$) & $w_{i}^{(t)} + \eta P_{i}y^{*}\odot\sigma'_{i}(z_{i})$ \\
     tpSGD\_{$\ell1$} & $||P_{i}y^{*} - y_{i}||_{1}$  & $w_{i}^{(t)} + \eta \text{sign}(P_{i}y^{*} - y_{i})\odot\sigma'_{i}(z_{i})$ \\
     tpSGD\_{$\ell2$} & $||P_{i}y^{*} - y_{i}||_{2}^{2}$ & $w_{i}^{(t)} + \eta (P_{i}y^{*} - y_{i})\odot\sigma'_{i}(z_{i})$ \\
    \hline
  \end{tabular}
    \caption{We show the relation between various projection-based methods for training neural networks. In particular, we look at the loss functions $J_{i}(.)$ that each approach is attempting to minimize as well as the weight update $w_{i}^{(t+1)}$ at layer $i$ for iteration $t+1$. Note that we incorporate multiplicative constants into $\eta$.}\label{tab:connections}
\end{table*}

tpSGD shares close connections with a few prominent BP-free, projection-based training algorithms. In some cases, tpSGD acts as a generalization. We highlight connections between Direct Feedback Alignment (DFA) \cite{nokland2016direct}, Error-Sign-Based Direct Feedback Alignment (sDFA) \cite{Frenkel_2021}, Direct Random Target Projection (DRTP) \cite{Frenkel_2021}, and our novel tpSGD approach (tpSGD using layer-wise $\ell1$-error (tpSGD\_{$\ell1$}) and $\ell2$-error (tpSGD\_{$\ell2$})).

Consider the case of a simple linear layer $i$ with non-linear activation function: $y_i = \sigma_{i}(z_i)=\sigma_{i}(x_iW_i)$. We define the following quantities, $J(.)$ is a loss function, $x_i$ is the input of layer $i$, $y_i$ is the output of layer $i$, $W_{i}$ are the weights associated with layer $i$, $\sigma_{i}(.)$ is the non-linear activation function associated with layer $i$, $\delta y_{i}$ is the estimated loss gradients for the outputs of layer $i$, $y_N$ is the predicted output of the final layer, $y^{*}$ is the ground truth one-hot encoding of the labels, $\sigma_{i}'(.)$ is derivative of the non-linear activation function of layer $i$, $P_{i}$ is the layer-dependent projection matrix of the targets for layer $i$, and $\eta$ is the learning rate.

In Table \ref{tab:connections}, we inspect the optimization objectives of the aforementioned algorithms and investigate how this affects the weight update steps at layer $i$. The layer-wise weight update rules are  similar to one another, but each algorithm has subtle distinctions in their assumptions, leading to different optimization behaviors and flexibility in the types of training pipelines they are compatible with. 

DFA, sDFA, and DRTP all attempt to minimize a global objective (cross-entropy between the ground truth labels and final layer output) using noisy gradient steps. DFA and sDFA require a complete forward pass through the network. Then these approaches compute estimated loss gradients for a given layer based on random projections of the gradients of the loss at the final layer. sDFA uses only the sign of the loss gradients whereas DFA uses the full magnitude of the loss gradients. Since both of the methods require a full pass through the network, they are not compatible with tpSGD.

DRTP makes use of the fact that labels are one-hot encodings to simplify sDFA, where DRTP's loss gradients are a surrogate of those of sDFA with shift and rescaling operations applied. Following the reasoning of Frenkel et al. \cite{Frenkel_2021}, we look at the behavior of the loss gradient $e_{sdfa} = sign(y^{*}-y_{N})$ when $y^{*}_{c}$ is 1 and when $y^{*}_{c}$ is 0. We assume $y^{*}_{c} \in \{0,1\}$ (i.e., the ground truth label for a given class is exactly zero or one) and $y_{Nc} \in (0,1)$ (i.e., the prediction of the final layer for a given class is in the range zero or one, non-inclusive). Under this assumption, $y^{*}_{c}$ never exactly equals $y_{N_c}$ (i.e., there is always some prediction error). The loss gradient for sDFA can be expressed as:
$$
e_{sdfa_c} = \text{sign}(y^{*}_{c}-y_{N_c}) =
\begin{cases}
1 & \text{if} \; c=c^{*} \\
-1 & \text{otherwise}
\end{cases} \\
,y^{*}_{c} \neq y_{N_c}
$$
$c$ is the index of the one-hot vector and $c^{*}$ is the true class.

The key observation here is that $e_{sdfa_c}$ will always be $+1$ when $y^{*}=1$, and it will always be $-1$ when $y^{*}=0$, regardless of the values of the prediction $y_{N}$, effectively decoupling the estimated loss gradients from the final layer predictions. The loss gradient associated with DRTP is simply a scaling and shifting of $e_{sdfa_c}$ under the above assumptions:
$$
e_{drtp_c} = \frac{1 + \text{sign}(y^{*}_{c}-y_{N_c})}{2} =
\begin{cases}
1 & \text{if} \; c=c^{*} \\
0 & \text{otherwise}
\end{cases} \\
,y^{*}_{c} \neq y_{N_c}
$$
Note $e_{drtp} = y^{*}$, meaning the one-hot labels can serve as the estimated gradient of the loss. To obtain layer-wise gradients w.r.t. the global loss, DRTP uses a random projection matrix to project these one-hot labels as the estimated loss gradients for the global objective. Frenkel et al. showed that for networks consisting of an arbitrary number of linear layers followed by a non-linear activation with cross-entropy as the loss function, DRTP produces estimated gradients within $90^{\circ}$ of the BP gradients, leading to learning, and they experimentally validated that DRTP works with less constrained feedforward architectures. DRTP does not require a full forward pass before updating the weights (no update locking), and DRTP can update layers independent of one another (no issues with weight transport), making DRTP fully compatible with tpSGD. Different layers in the same network can be optimized via either tpSGD or DRTP without issue. In the rest of the paper, we regard DRTP as a subset of the capabilities provided by tpSGD. When DRTP is mentioned, we specifically refer to the aforementioned variations of cost functions and weight updates.   

Finally, we consider our novel tpSGD-style learning. In contrast to other approaches, tpSGD tries to minimize a collection of local losses, with the intention of learning representations that promote reduction of the global loss. tpSGD locally optimizes layers, forcing the layer outputs to align with random projections of the one-hot ground truth labels (serving as layer-wise targets for optimization). tpSGD tries to learn representations such that instances from the same class are well-separated from instances of other classes within the same layer, and representations from two different layers should be well-separated from one another. tpSGD is compatible with any generic layer-wise loss between the layer-wise targets and layer-wise outputs, but we explicitly consider the $\ell1$-norm and $\ell2$-norm for this paper.

\subsection{Layer Projection Definitions}
In tpSGD, we provide learning targets for intermediate layers by projecting the ground truth labels via randomly generated matrices. This section discusses the projection strategies for linear, convolutional and recurrent layers in tpSGD.

\subsubsection{Linear layer projections}

In the simplest case of a fully connected layer with $n$-nodes and a classification problem with $bs$ batch size, $nc$ classes, we map a batch of labels with dimensions $(bs, nc)$ to one with $(bs, n)$ by multiplying by a random matrix $P$ with dimensions $(nc, n)$.  Among our experiments we tested sampling from different distribution functions (uniform, normal, ...),  but have found no noticeable difference on the performance.

\subsubsection{Conv2D layer projections}
To extend target projection via random matrices for CNN, we implemented two approaches to generate the target features for these intermediate Conv2D layers, which are illustrated in Figure \ref{fig:conv_projection} a) and b).  

Assume the output of a given Conv2D layer has dimensions $(bs, nx, ny, nf)$, where $bs$ is the size of the mini-batch being processed, $nx$ and $ny$ are the $x$ and $y$ dimensions of the filtered image, and $nf$ are the number of filters. In panel a), the naive approach, we generate a single, long projection matrix $P$ with dimensions $(nc, nx\times ny\times nf)$ and reshape the output to match the target dimensions. In panel b), we propose a filter-based sampling approach: we generate $nf$ different projection matrices $P_i$ of size $(bs, nx\times ny)$ for each of the $i \in \{1, 2, ..., nf\}$ filters, sampling from a different random distribution for each.  We sample from normal distributions with varying standard deviations in the range $[0, 1)$ at equally spaced intervals $\sigma_i = i/nf$. 

\begin{figure}
  \centering
  \includegraphics[width=0.85\columnwidth]{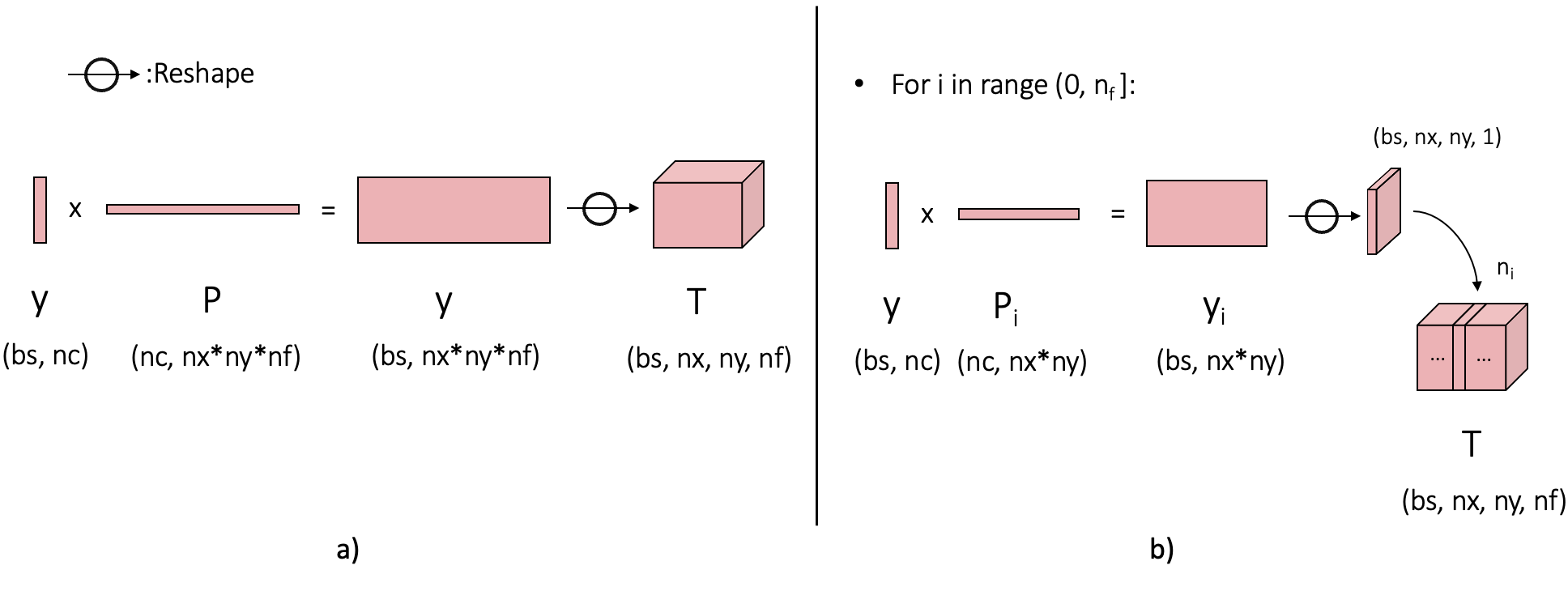}
  \caption{Illustration showing the difference between approaches to constructing target features T for convolutional layers using random matrices P.  Panel a) shows the basic or "naive" approach vs. filter-based sampling.  In b), the tuples in parenthesis under the labels illustrate the tensor sizes.}\label{fig:conv_projection}
\end{figure}

\begin{figure}
  \centering
  \includegraphics[width=0.4\columnwidth]{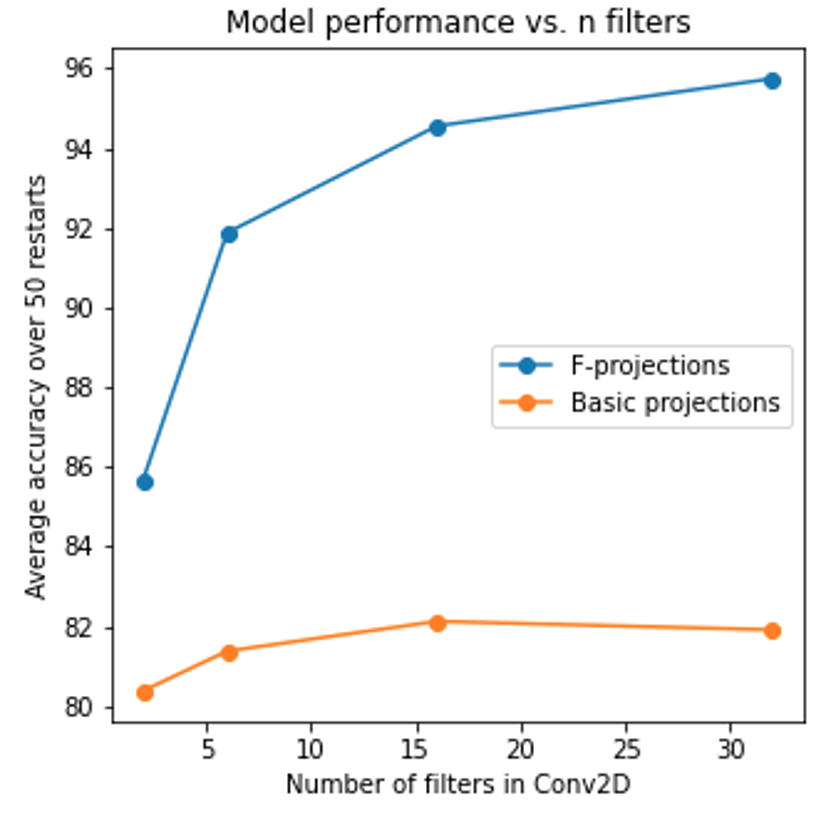}
  \caption{Basic and filter-based projection (F-projections) performance for Conv2D layers as a function of the number of filters.}\label{fig:filter_sampling}
\end{figure}

 
 Figure \ref{fig:filter_sampling} shows the difference in performance between these two sampling methods.  A model consisting of a Conv2D layer, leaky ReLU activation and a final linear classification layer was trained using our tpSGD algorithm on MNIST using both the basic sampling and proposed filter-based sampling method. Whereas the basic sampling shows little or no improvement as the number of filters in the layer increases, filter-based sampling projections show steady improvement, albeit with diminishing returns.  In our studies we observed a small, constant increase in training time due to the increased cost in initializing the projections for each Conv2D layer.

\subsubsection{Recurrent layer projections}
Past work \cite{Frenkel_2021} showed that target projection can train feedforward networks. To the best of our knowledge, target projection has not previously been shown to work with recurrent neural networks (RNNs). We describe how recurrent layers can be modeled so that simple modifications of either DRTP and tpSGD makes target projection an effective algorithm for training RNNs. In subsequent sections, we demonstrate experimentally for the first time that random target projection promotes learning in recurrent networks.


We formulate a simple recurrent cell (pictured in Appendix \ref{app:unrolled}), which computes the following:
$$H_{t+1} = \text{sigmoid}(H_{t}W_{H} + b_{H} + X_{t}W_{X} + b_{X})$$
where $H_{t}$ is the hidden state input at time $t$, $X_{t}$ are the input features at time $t$, $W_{H}$, $b_{H}$, $W_{X}$, and $b_{X}$ are the learnable parameters of the cell (shared over all time steps), and $H_{t+1}$ is the hidden state that serves as input to the next time step. While there are more complex cell types used in practice (e.g., LSTM \cite{hochreiter1997long} and GRU \cite{cho2014properties} cells), these are designed to solve issues with modeling long-range dependencies (e.g., vanishing gradients), which random projection-based approaches are not susceptible to, so we do not consider them for this paper.

In order to apply TP to this recurrent cell, similar to BP through time \cite{werbos1990backpropagation}, we \emph{unroll} the recurrent cell over time for a fixed problem-specific sequence length (see Appendix \ref{app:unrolled}). This produces a feedforward-like network where every layer shares a common set of learnable parameters but uses the hidden state output by the previous layer with time step-dependent input features.


We see two key differences between the unrolled recurrent layer and generic feedforward MLPs: 1) there are two linear functions that feed into the non-linearity per unrolled ``layer'', and 2) the weights are shared between the unrolled ``layers''. The first difference is handled by updating the parameters of each function separately. For (2), we can make use of the fact that we can freeze the weights of the recurrent cell for the current training iteration and update over all time steps simultaneously, i.e., with frozen weights, a forward pass through the unrolled recurrent cell is:

\begin{equation}
\begin{split}
\begin{bmatrix}
H_{1} \\ H_{2} \\ ... \\ H_{end}
\end{bmatrix}
& = \sigma
\Biggl(
\begin{bmatrix}
H_{0}X_{0} \\ H_{1}X_{1} \\ ... \\ H_{N}X_{N}
\end{bmatrix}
\begin{bmatrix}
W_{H} \\
W_{X}
\end{bmatrix}
\oplus 
(b_{H} + b_{X})
\Biggr)
\end{split}
\end{equation}
where $\oplus$ represents row-wise addition of a vector to a matrix. Since these weights are frozen and updated simultaneously for all time steps and tpSGD doesn't enforce any backward propagation requirements, the $H$s can be treated as independent inputs. Then, the above formulation of the recurrent layer is equivalent to a feedforward linear layer plus nonlinear activation.

When computing the gradient of $W_{H}$, there is no dependency on $X$ and when computing the gradients of $W_{X}$, there is no dependency on $H$, so each set of parameters can be updated separately: when $W_{H}$ is updated $W_{X}$ is frozen and vice versa. Secondly, when gradients are computed using this formulation, it is equivalent to computing gradients over each ($H_t$, $X_t$) separately and summing over all time steps (i.e., gradient accumulation). This means we can compute the individual pseudo-gradients per time step and sum or average them over all time steps, \textit{storing only the accumulated gradients, reducing memory requirements}. To train the parameters of a recurrent cell for a given time step, we use the following updates, where we assume different projection matrices $B_t$ for the optimal one-hot encoded labels $Y^*$ for each time step with learning rate $\eta$ for tpSGD\_$\ell1$:

\begin{equation}
\small
\begin{split}
&\begin{bmatrix}
W_{H} \\ b_{H}
\end{bmatrix}^{(k+1)}
=
\begin{bmatrix}
W_{H} \\ b_{H}
\end{bmatrix}^{(k)}
+
\eta
\begin{bmatrix}
\nabla W_{H} \\ \nabla b_{H}
\end{bmatrix}^{(k)}
\end{split}
\end{equation}

\begin{equation}
\small
\begin{split}
&\begin{bmatrix}
W_{X} \\ b_{X}
\end{bmatrix}^{(k+1)}
=
\begin{bmatrix}
W_{X} \\ b_{X}
\end{bmatrix}^{(k)}
+
\eta
\begin{bmatrix}
\nabla W_{X} \\ \nabla b_{X}
\end{bmatrix}^{(k)}
\end{split}
\end{equation}

\begin{equation}
\small
\begin{split}
\begin{bmatrix}
\nabla W_{H} \\ \nabla b_{H}
\end{bmatrix}
=
\sum_{t=0}^{N}
\begin{bmatrix}
 H^T_{t}\\1^T
\end{bmatrix}
D_H, 
\begin{bmatrix}
\nabla W_{X} \\ \nabla b_{X}
\end{bmatrix}
=
\sum_{t=0}^{N}
\begin{bmatrix}
 X^T_{t}\\1^T
\end{bmatrix}
D_H
\end{split}
\end{equation}


\begin{equation}
 D_H = \text{sign}(Y^*B_t - H_{t+1}) \odot H_{t+1} \odot (1-H_{t+1})   
\end{equation}
where $\odot$ is element-wise multiplication and ``1'' represents the generalized scalar/vector/matrix of all ones.

We can similarly derive update rules for tpSGD\_$\ell2$ and DRTP. We found experimentally that tpSGD\_$\ell2$ does demonstrate non-trivial training of RNNs, but it significantly underperforms tpSGD\_$\ell1$ and DRTP in this case.

The initial weights of the recurrent cells can be all zeros or drawn uniformly randomly with small magnitude. To initialize the projection matrices for tpSGD\_$\ell1$, we use an random binary matrices with approximately orthogonal rows. To initialize the projection matrices for DRTP, we use random matrices with orthogonal rows.

Weights can be updated locally using multiple forward passes through the recurrent layer before sending the final $H_{end}$ feature vector to the next tpSGD-compatible layer. This means it is trivially easy to extend the system to stacked RNNs \cite{hihi1995hierarchical} and with simple modifications, the recurrent cell can be made to be compatible with bidirectional RNNs \cite{schuster1997bidirectional}.
  
\section{Experiments and Results}

We compare tpSGD to BP over a set of classification tasks and different datasets, and combinations of Linear, Conv2D, and Recurrent layers to study how effectively the algorithm can scale. We also compare it to a series of backpropagation-free algorithms designed specifically to work at the edge. Both ZORB and LM-KARnet ingest an entire dataset in a single step during optimization. They excel in training with small datasets, but fail to extend to large datasets, deep NNs and complex tasks.  
 
\subsection{Shallow networks, MNIST and CIFAR}

We begin by studying shallow networks on MNIST \cite{deng2012mnist} and CIFAR \cite{Krizhevsky09learningmultiple}.    Table \ref{tab:mnistgradfree} shows a comparison between the proposed tpSGD, ZORB and LM-KARNet on MNIST using a 2-layer MLP with leaky ReLU activation. We report  mean value and standard deviation in parenthesis over 25 random restarts for the training time and model accuracy.  All three algorithms were benchmarked using implementations in TensorFlow 2.9 running on a single NVIDIA RTX A5000 GPU. While tpSGD's performance is 1-2\% below both ZORB and LM-KARNet, we see sizeable time savings for tpSGD. While both ZORB and LM-KARnet use the MP inverse to obtain the new weights, LM-KARnet replaces target propagation in ZORB with random projection. This accounts for the substantial speed up when comparing their training times: ZORB inverts each layer starting from the output while LM-KARnet projects using a single random matrix. However, LM-KARnet still uses the MP inverse to optimize each layer's weights. Replacing this expensive operation in tpSGD leads to further decreases in the training time with near negligible loss in accuracy.

\begin{table}
  \label{sample-table}
  \centering
  \begin{tabular}{|c|c|c|}
    \hline
    \multirow{2}{*}{}
     Algorithm     & MNIST   & Training  \\
          & Test accuracy (\%)     & Time (s)  \\
    \hline
    ZORB & 89.4(0.22)  & 7.9(0.39)     \\
    LM-KARNet & 89.9(0.12) & 4.9(0.38)     \\
    tpSGD  & 88.3(0.25) & 1.7(0.01) \\
    \hline
  \end{tabular}
    \caption{Comparison of gradient free approaches with a 2-layer MLP trained on MNIST}\label{tab:mnistgradfree}
\end{table}

In Table \ref{tab:mnist_cifar} we compare the performance (mean and standard deviation per metric) of a shallow model consisting of two Conv2D layers with leaky ReLU activations, and one final Linear layer on MNIST and CIFAR-10 using the same settings described earlier. The results show that tpSGD can nearly match the accuracy (within 1\%) and training time with BP.  Appendix \ref{app:tpsgd_scaling} explores the scaling of tpSGD (as compared to BP) in more detail in CNNs with up to 7 trainable (Dense and Conv2D) layers.

  \begin{table}
  \label{sample-table}
  \centering
  \begin{tabular}{|c|c|c|c|c|}
    \hline
    & \multicolumn{2}{|c|}{tpSGD} & \multicolumn{2}{|c|}{BP} \\
    \hline
        & Acc.(\%) & Time(s) & Acc.(\%) & Time(s) \\
    \hline
    MNIST & 96.6(0.2) & 2.7(0.1) & 97.1(0.21) & 2.3(0.3) \\
    CIFAR & 46.5(1.2) & 4.6(0.1) & 46.8(1.1) & 4.8(0.4) \\
    \hline
  \end{tabular}
  \caption{Comparison between tpSGD and BP on training a shallow CNN on MNIST and CIFAR-10}\label{tab:mnist_cifar}
\end{table}



\subsection{VGG and Imagenette}
  
  Next, we experiment extending tpSGD to large images and deeper networks.  For this we employ Imagenette \cite{imagewang}, a subset of 10 easy classes from Imagenet \cite{imagenet_cvpr09}.  Imagenette consists of roughly 1000 color images, at least 320x320 pixels in size. The first row in Table \ref{tab:imagenette_results} compares the performance between VGG11 networks (with 11 trainable layers) trained from a random initialization using BP and tpSGD.  We report mean and variation (in parenthesis) of the performance over 25 random restarts. We present the first practical demonstration of backpropagation-free training applied to images and networks of this scale, the SoA results obtained by tpSGD (~65\%) is roughly 10-15\% drop in performance compared with the BP baseline. 
  
  Given the challenge of directly scaling-up tpSGD, we studied an alternative approach to scaling to DNNs by adapting to domain-shift with shallow adaptor NNs running tpSGD. The concept is to design and optimize a few adaptor layers that connect to the layers of the pretrained DNNs. During learning, only the shallow adaptor weights are trained and the pretrained DNNs are fixed. Different from the SoA finetuning methods, the configurations of the adaptors at the edge are determined via network architecture search \cite{mellor2021neural} and optimized based on out-of-distribution detection \cite{wilson2021hyperdimensional}. The discussion of these is not the focus of this paper. The second and third rows in Table \ref{tab:imagenette_results} show studies using tpSGD and BP in a transfer learning setting.  The second row shows the baseline approach using BP to finetune the final fully connected layers in the network.  To compare, the third row shows the performance obtained using tpSGD and BP using a lightweight adaptor configuration obtained via a basic network architecture search (NAS) (detail discussed in appendix \ref{app:adaptorsearch}).  The adaptor consists of: Conv2D(filters=64, kernel size=5, strides=1), LeakyReLU, Dense(256), LeakyReLU, Dense(n classes=10).

  The model trained using tpSGD on adaptors is within 3-5\% of the result obtained using BP and Adam on VGG16 finetuning echoing the results shown in earlier sections discussing shallow networks, while significantly reducing the in memory footprint due to the selected adaptor architecture. We envision leveraging a pretrained and quantized or pruned feature extractor on specialized hardware together with tpSGD-based shallow adaptors to enable training on edge devices after deployment and adapting to new domains, novel tasks and distribution shifts.
  
  \begin{table}
  \label{sample-table}
  \centering
  \begin{tabular}{|l|cc|}
    \hline
    Experiment  & \multicolumn{2}{|c|}{Imagenette Accuracy (\%)}\\
        & tpSGD & BP \\
    \hline

    Full net (180 MB)  &  64(3.1)     & 78(1.4)  \\
    \hline
    Transfer (492 MB)& - & 87(1.3) \\
    \hline
    Adaptor (6 MB) & 83(0.95) & 88(1.7) \\
    \hline
  \end{tabular}
  \caption{Performance comparison between tpSGD and SGD BP on the Imagenette dataset in three secenarios: training all layers of a VGG11, the transfer learning baseline and using our own adaptor with a fixed pretrained VGG16.  We include, in MB, the total size of layers being trained.}\label{tab:imagenette_results}
\end{table}

\begin{table*}[!ht]
\resizebox{1.0\textwidth}{!}{
    \centering
    \begin{tabular}{|c|c|c|c|c|c|}
    \hline
    \multirow{2}{*}{}
        Dataset & Random & Random RNN + & Trained RNN (DRTP) & Trained RNN (tpSGD\_$\ell1$) + & Trained RNN (BP) + \\
          & & Trained Classifier (tpSGD\_$\ell2$) & Classifier (tpSGD\_$\ell2$) & Classifier (tpSGD\_$\ell2$) & Classifier (BP) \\ \hline
   
        Seq. MNIST (rows) & 9.58\% (10\%) & 9.76\% & 83.39\% & 79.81\% & 92.53\% \\ \hline
        Seq. MNIST (pixels) & 10.28\% (10\%) & 11.35\% & 62.83\% & 66.46\% & 75.55\% \\ \hline
        EEG Seizure & 48.35\% (50\%) & 79.48\% & 89.57\% & 83.57\% & 94.14\% \\ \hline
        UCF101  & 0.80\% (0.99\%) & 53.78\% & 60.18\% & 60.76\% & 61.58\% \\ \hline
        Twitter (one-hot)  & 49.92\% (50\%) & 50.08\% & 66.82\% & 67.72\% & 71.75\% \\ \hline
        Twitter (Word2Vec) & 49.92\% (50\%) & 64.84\% & 69.60\% & 68.67\% & 72.26\%  \\ \hline
    \end{tabular}
    }
\caption{Results of training a single-layer RNN with linear classifier. Numbers in parentheses are true random chance. We show the algorithm used to train each layer in parentheses where BP is traditional backpropagation.}
\label{tab:rnn_basic}
\end{table*}

\subsection{Time Series Analysis Using Recurrent Neural Networks}
\label{sec:rnn_experiments}
We also conducted experiments over a wide range of datasets with differing properties between them (see Table \ref{tab:rnn_datasets} in Appendix \ref{app:rnn_datasets}), to validate the training of the recurrent layers in combination with multi-layered perceptrons for classification of time series data. In the case of the UCF101 and ``Twitter Sentiment Analysis - Word2Vec Encoding'' datasets, we use an external model to first preprocess the data into feature vectors. All other datasets work over the raw representations.

We consider three experiments: (1) a single-layer RNN in combination with a linear classifier in Table \ref{tab:rnn_basic}, (2) a two-layer stacked RNN in combination with a two-layered perceptron (Table \ref{tab:rnn_stacked} of Appendix \ref{app:rnnscaling}), and (3) a single-layer Bidirectional RNN in combination with a linear classifier (Table \ref{tab:rnn_bidirectional} of Appendix \ref{app:rnnscaling}). In each case, we use a hidden vector size of 512, a batch size of 128, and train for 20 epochs. We compare against random chance, a completely random network of equivalent structure, a network where the RNN is random, but the classifier is trained, and a structurally-equivalent model trained only using backpropagation. We also demonstrate that the method effectively trains with either DRTP or tpSGD\_{$\ell1$} as the optimizer for the recurrent layers, and performance is generally similar.

We verify that there is non-trivial learning of the recurrent layer(s) by comparing the model with randomly weighted recurrent layers + trained classifier with the model where both the recurrent layers and classifiers are trained using target projection. We see significant improvement in all cases except the Seizure Activity Recognition dataset where even random features achieve relatively high performance. 

We also compare performance of the target projection-based models to the performance of the gold-standard BP based models to understand how well the model learns compared to an ``upper bound'' and similarly, compare to random chance and models with random weights to and how well the model learns compared to a ``lower bound''. In all cases, the trained model significantly outperforms the completely random models. Generally, the models trained with target projection perform well. The target projection-based model performs within 5\% accuracy of the backpropagation based model on 3 ($tpSGD_{\ell1}$) / 4 (DRTP) of the 6 datasets. A larger gap exists when tested on both forms of sequential MNIST, but the model still exhibits significant performance.

Similar results are seen for the stacked and bidirectional RNNs. One interesting observation about these models is that performance is very similar to the basic single layer RNN with linear classifier model. This suggests that target projection is not preventing the stacked and bidirectional RNNs from learning meaningful representations. 

\section{Conclusions and Future Work}
  In this paper, we explored a novel BP-free, feedforward-only optimization algorithm designed to enable training in resource constrained environments, such as edge devices.  We discussed the connections between tpSGD and other existing BP-free algorithms and compared their performance when training in architectures such as MLPs, CNNs and RNNs. We found that tpSGD in training performs comparably to the BP SGD and BP-free algorithms in shallow MLPs, CNNs, and RNNs, and is superior to other BP-free algorithms in terms of memory and time.  Finally, we showed tpSGD scales-up to DNNs using transfer learning via an optimized shallow adaptor concept. Although  tpSGD-based training of full DNNs underperforms compared to BP SGD, the algorithms are effective for training shallow adaptors connected to fixed encoders. This alternative scale-up architecture has great potential for on-device learning in edge scenarios in order to tackle domain shift, novel tasks, and other similarly challenging problems.
  
  For future work, we desire to better understand tpSGD from a theoretical perspective. We have noticed that tpSGD-based models do not seem to achieve their full potential in DNNs. We hypothesize that this may be due to (the combination of) two reasons: (1) more complex models might not be needed for relatively simple benchmark datasets; (2) because the layers of the model are trained without feedback from later to earlier layers, these more complex models might be overly restricted in what they can learn, thus showing minimal growth in performance. 
  
  From the filter-based sampling in CNN, we have noticed that carefully selecting the distinctive projection distribution is important. In the future, we want to explore whether we can learn the distribution of layer-wise target projections from pretrained models for the dataset instead of blindly estimating these projections. Another future direction is to analyze eigenvector correlations among input data samples based on per layer Jacobians \cite{mellor2021neural} so that the random projection is selected with the least redundancy among the estimated projections.          

\section*{Acknowledgements}
This research is based upon work supported in part by the Office of the Director of National Intelligence (ODNI), Intelligence Advanced Research Projects Activity (IARPA), via Contract No: 2022-21100600001. The views and conclusions contained herein are those of the authors and should not be interpreted as necessarily representing the official policies, either expressed or implied, of ODNI, IARPA, or the U.S. Government. The U.S. Government is authorized to reproduce and distribute reprints for governmental purposes notwithstanding any copyright annotation therein.

\bibliographystyle{ieeetr}
\bibliography{bibliography.bib}

\appendix
\onecolumn
\section{CNN Scaling Studies}\label{app:tpsgd_scaling}

\begin{figure}
  \centering
  \includegraphics[width=.7\columnwidth]{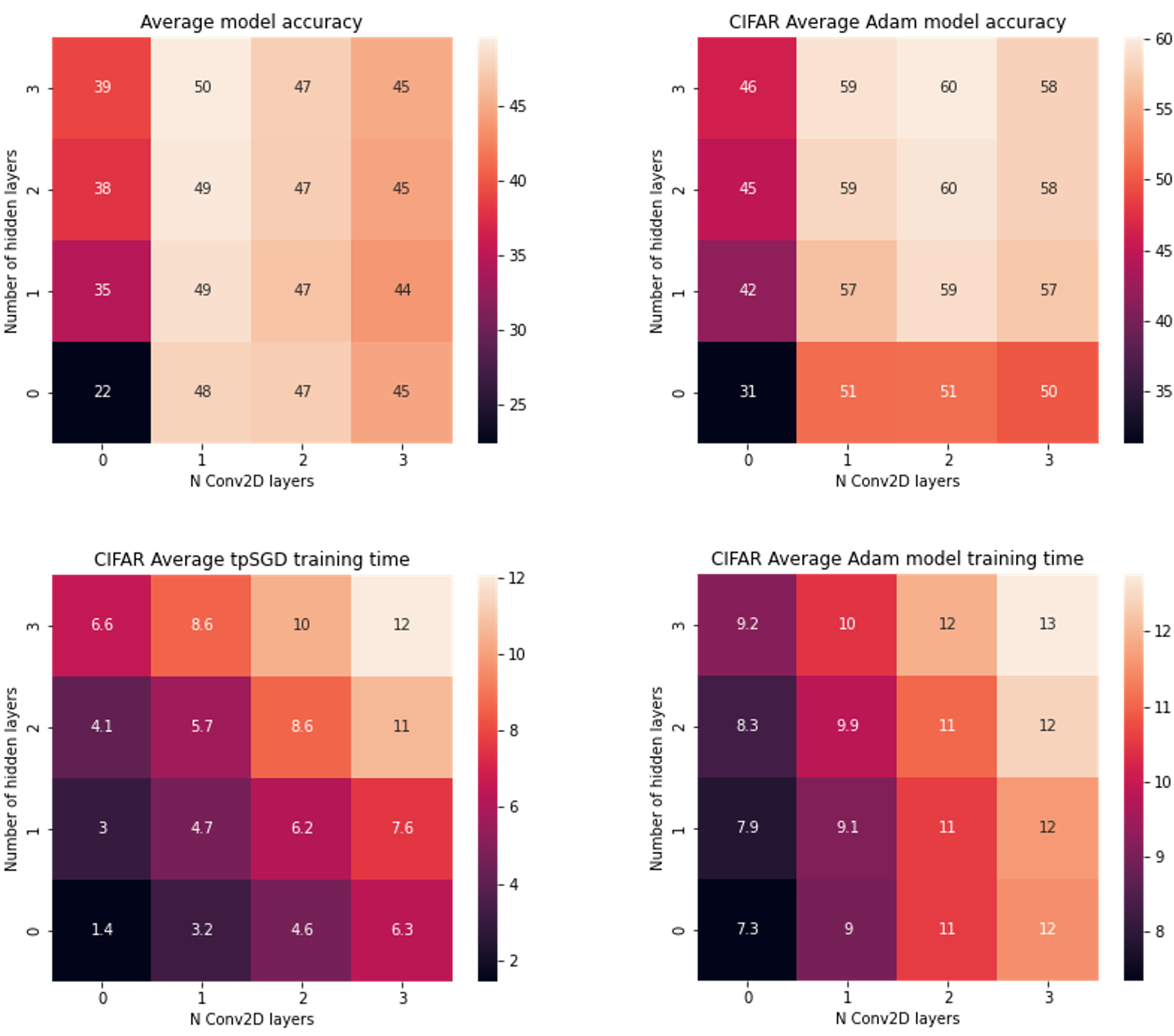}
  \caption{Heatmap showing model performance (top) and training time (bottom) on CIFAR as a function the number of hidden linear ($y$-axis) and Conv2D ($x$-axis) layers using tpSGD (left) and Adam (right) benchmark.  All hidden layers have 256 nodes, while all Conv2Ds have kernel size=(5,5),  filters=16, stride=(1,1).  All models are followed by a final, fully connected layer mapping to the classification space.}\label{fig:cifarnas}
\end{figure}

To study the performance and feasibility for scaling to deeper models, we perform a basic grid search over different network configurations, and study the performance of these using tpSGD and Adam on CIFAR.  These results are summarized in Figure \ref{fig:cifarnas}. The left diagrams, or heat maps, correspond to the results obtained using tpSGD while the results from Adam are shown on the right.  The top row displays the average model performance over 25 random restarts while the bottom row shows the average training time.

Moving along the $x$-axis of any of the heat maps corresponds to adding a Conv2D layer with kernel size 5, stride 1 and leaky ReLU activation at the beginning of the network.  Moving up the $y$-axis  corresponds to adding Linear layers with 256 hidden nodes and leaky ReLU activation before the final classification layer.  As such the smallest model shown (bottom left) has a single trainable layer, and the largest has 10.

Appendix \ref{app:tpsgd_scaling} shows a detailed scaling study comparing models trained with tpSGD as compared to backpropagation.  These results show that the proposed filter-based sampling approach for generating random projections is close to capturing the expressive power of CNNs trained using Adam.  In particular for Number-of-hidden-layers (Nhidden) = 0, the scaling as a function of nconv is within 3-5\% to the baseline across the networks studied.  However, looking at the upper right corner, at the deepest networks, the difference in performance is much larger (10-15\%), as adding Linear layers in tpSGD is providing less of an improvement relative to the Adam baseline. We will discuss ideas of better random projections for linear layers in the future work section. 

\section{VGG16 Adaptor Configuration}\label{app:adaptorsearch}
\begin{figure}
  \centering
  \includegraphics[width=.8\columnwidth]{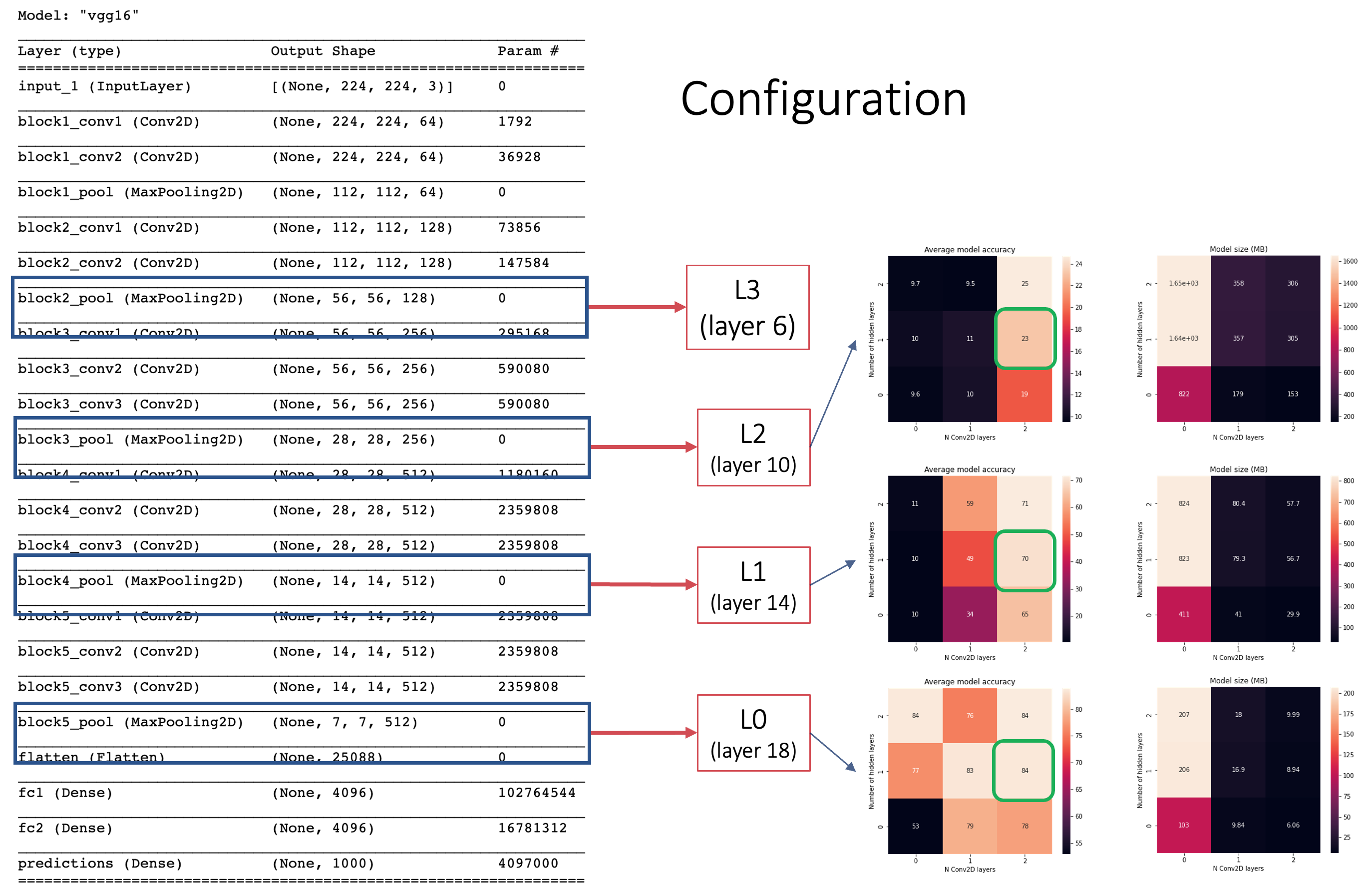}
  \caption{Schematic showing the naive adaptor search and relevant studies for VGG16.  The left represents the feature extractor and the rightmost heat maps show the model performance and size for different configurations.}\label{fig:adaptor_search}
\end{figure}

  This section discusses briefly the adaptor definition and related studies used to benchamrk tpSGD against BP in transfer learning or domain adaptation setting.   Diagram \ref{fig:adaptor_search} schematically illustrates the procedure.  The backbone on the left is the feature extractor portion of VGG16, and the red arrows illustrate "tap" points where we extract intermediate representations, which can then be used to train models on the classification task.
  
  The heatmaps associated to each adaptor level $L_i$ show the model performance (left) and memory footprint as we increase the number of convolutional layers (x axis) or hidden, dense layers (y axis). 
  The final selected configuration for each layer is shown in green, consisting in each case of the following:  Conv2D(filters=64, kernel size=5, strides=1), LeakyReLU, Dense(256), LeakyReLU, Dense(10).  

  \begin{figure}[H]
  \centering
  \includegraphics[width=0.5\columnwidth]{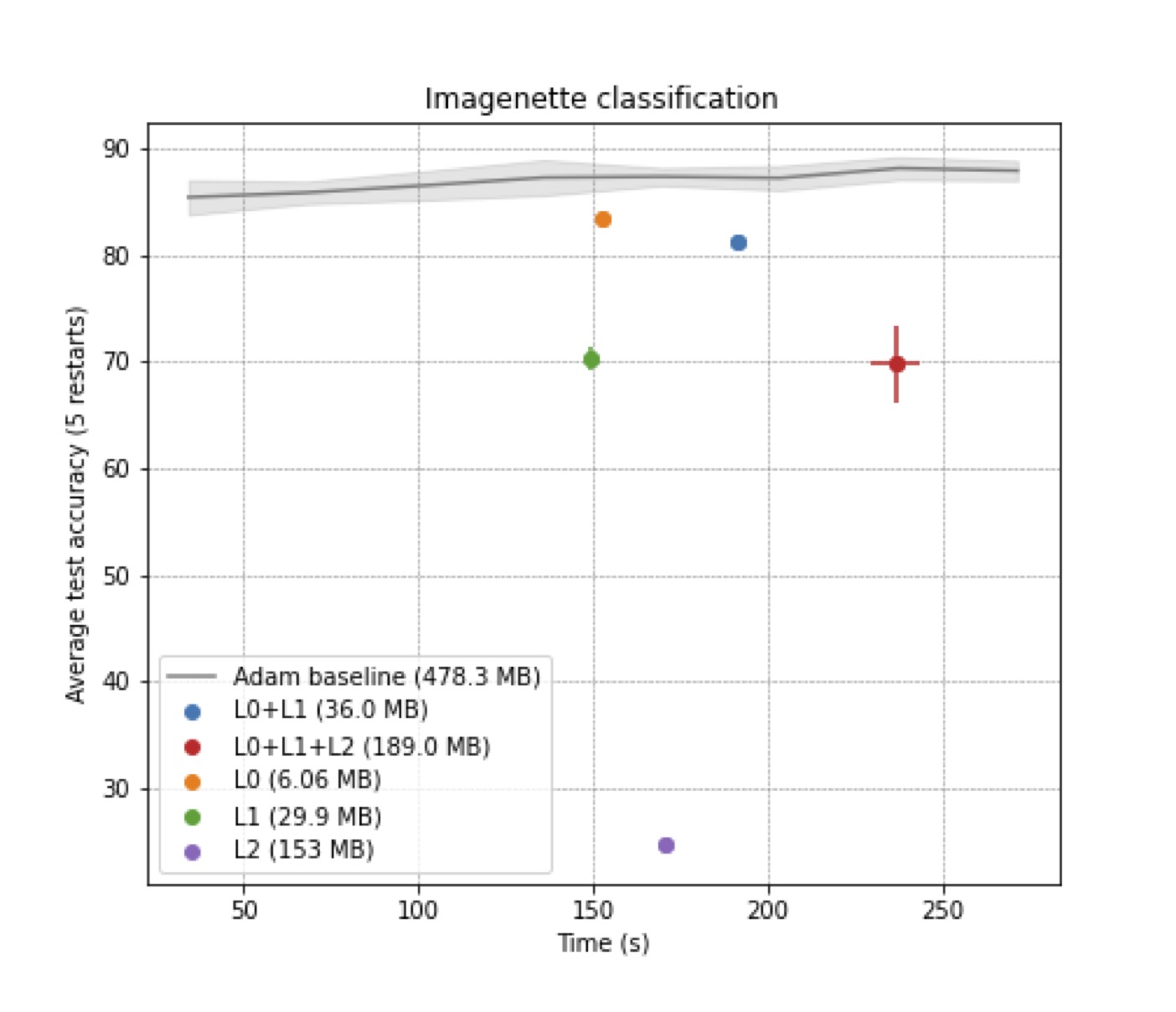}
  \caption{tpSGD adaptor performance on Imagenette using different adaptor configurations, compared to the BP transfer learning baseline.  The size of the trained models is show in parenthesis.}\label{fig:vgg166adaptors}
\end{figure}

  Figure \ref{fig:vgg166adaptors} shows a comparison between the performance of the different adaptors ($L_0, L_1, L_2$) and combinations of them, as well as the transfer learning baseline obtained by finetuning the final fully connected layers of VGG16 with BP.  The results with the single L0 adaptor are within 3\% of the BP result, with a substantial decrease in the memory footprint due to the selected network architecture.  It is interesting to note that for this specific dataset adding additional adaptors does not improve the performance. We have experimented with other datasets domain-transferring from RGB to the depth map using EfficientNet v2 at the edge. The optimized adaptors are selected by NAS to be connected to the 2 middle layers (block 4 and 5) of the network and surprisingly not to the last fully connected layer.

\section{Basic Unrolled Recurrent Cell}
\label{app:unrolled}
\begin{figure}[H]
  \centering
  \includegraphics[width=0.7\columnwidth]{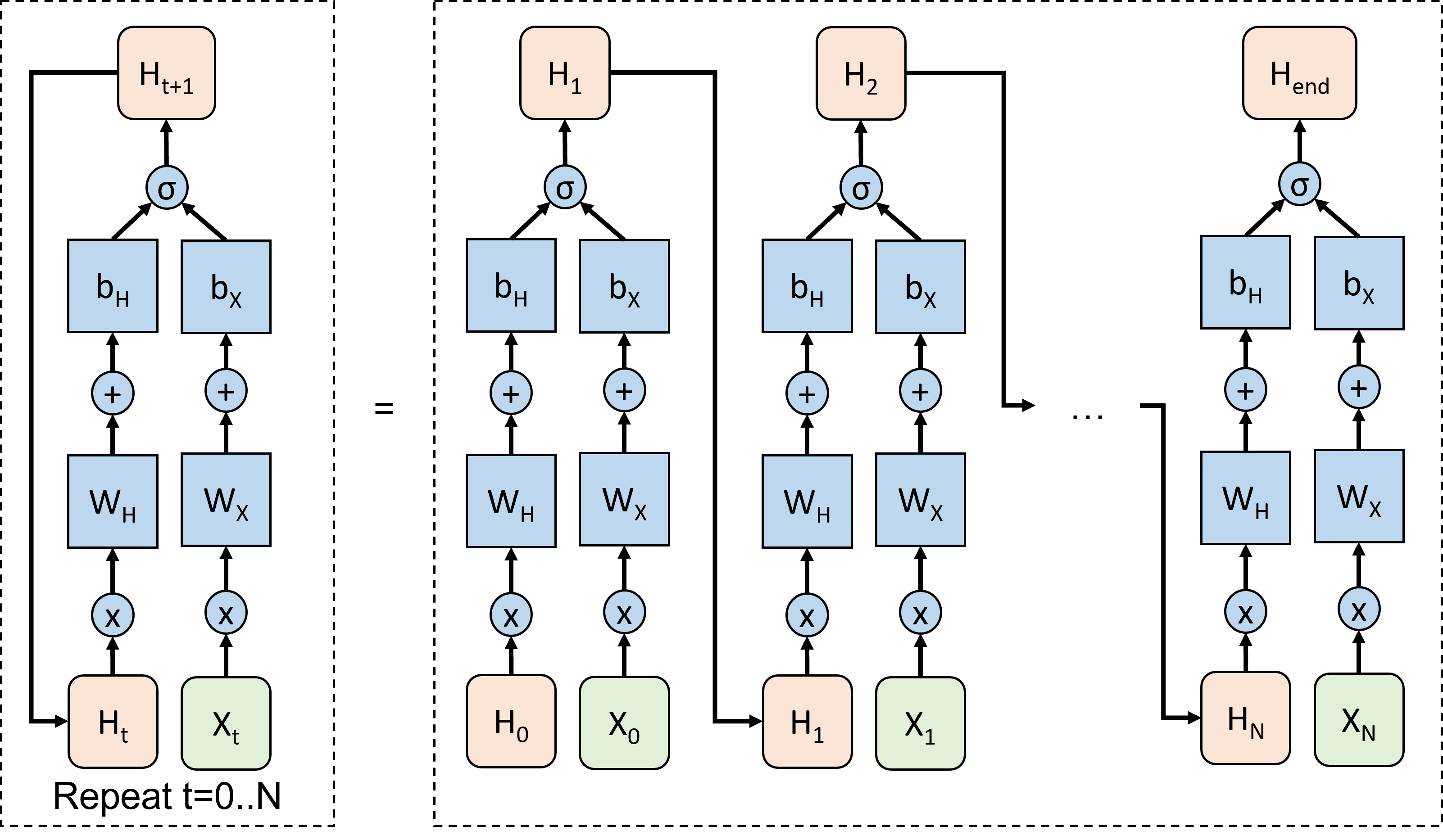}
  \caption{Formulation of the simple recurrent cell used in our system and its ``unrolled'' equivalent.}\label{fig:rnn_combined}
\end{figure}
In Figure \ref{fig:rnn_combined}, we show a pictorial representation of the unrolled recurrent cell used to formulate the basic tpSGD update rules for recurrent neural networks.
  
\section{RNN Dataset Properties}
\label{app:rnn_datasets}
  In this section we discuss in more detail the different datasets used to benchmark tpSGD on time series data using RNNs. Table \ref{tab:rnn_datasets} shows the general properties of six different benchmark datasets: Sequential MNIST (row- and pixel-wise), EEG seizure signals, UCF 101 activity detection dataset, and the Twitter Sentiment Analysis dataset (where input features are one-hot word vectors and wave2vec encodings). Note the wide range in dataset characteristics. These datasets include a mix of modalities (images, language, 1-d signals), different sequence lengths, variance in the number of features per sequence token, several different data types of the features with different ranges of values and amount of sparsity, varying amounts of training data, and a wide range in the number of labels for classification. Altogether, this set of benchmarks captures the ability of the proposed tpSGD method to generalize to diverse sequence classification problems. 
  
\begin{table*}[!htp]
\resizebox{1.0\textwidth}{!}{
    \centering
    \begin{tabular}{|c|c|c|c|c|c|c|c|c|c|}
    \hline
    \multirow{2}{*}{}
        Dataset & Modality & Number of  & Number of & Number of & Sequence & Number of & Feature & Sparsity & Range\\
         &  & Samples (Training) & Samples(Test) & Classes & Length & Features & Type & of Features & \\ \hline
    \multirow{2}{*}{}    
        Sequential MNIST & Grayscale Images & "60,000" & "10,000" & 10 & 28 & 28 & Float & 80.90\% & [0, 1] \\
        Rows \cite{lamb2016professor,deng2012mnist} & & & & & & & & & \\ \hline
    \multirow{2}{*}{}    
        Sequential MNIST & Grayscale Images & "60,000" & "10,000" & 10 & 49 & 1 & Float & 80.90\% & [0, 1] \\
        Pixels \cite{lamb2016professor,deng2012mnist} & & & & & & & & & \\ \hline
    \multirow{2}{*}{}    
        EEG Seizure & 1-D EEG Signal & "3,450" & "1,150" & 2 & 45 & 1 & Int & 0.00\% & [-1885, 2047] \\
         Recognition \cite{andrzejak2001indications} & & & & & & & & & \\ \hline
    \multirow{2}{*}{}
        UCF101 Action  & RGB Videos & "9,518" & "3,769" & 101 & 10 & 2048 & Float & 41.20\% & [0, 15.9375] \\
        Recognition \cite{soomro2012ucf101} & & & & & & & & & \\ \hline
    \multirow{2}{*}{}    
        Twitter Sentiment Analysis  & Natural Language & "75,000" & "25,000" & 2 & 15 & 1000 & Binary & 99.90\% & \{0, 1\} \\
        One-Hot Encoding \cite{go2009twitter} & & & & & & & & & \\ \hline
    \multirow{2}{*}{}
        Twitter Sentiment Analysis & Natural Language & "75,000" & "25,000" & 2 & 15 & 100 & Float & 54.30\% & [-3.35, 3.12] \\
        Word2Vec Encoding \cite{go2009twitter,mikolov2017advances} & & & & & & & & & \\ \hline
    \end{tabular}
    }
\caption{Properties of the datasets used for testing the capabilities of the recurrent cell}
\label{tab:rnn_datasets}
\end{table*}

\section{RNN Scaling Studies}\label{app:rnnscaling}
As mentioned in the main experiments section, we demonstrate how tpSGD can scale to more complex RNN architectures. We consider two such cases: multi-layer stacked RNNs and bidirectional RNNs. We report full results in Tables \ref{tab:rnn_stacked} and \ref{tab:rnn_bidirectional}, and we discuss findings in Section \ref{sec:rnn_experiments}.

\begin{table*}[!htp]
\resizebox{1.0\textwidth}{!}{
    \centering
    \begin{tabular}{|c|c|c|c|c|c|}
    \hline
    \multirow{2}{*}{}
        Dataset & Random & Random RNN + & Trained RNN (DRTP) & Trained RNN (tpSGD\_$\ell1$) + & Trained RNN (BP) + \\
          & & Trained Classifier (tpSGD\_$\ell2$) & Classifier (tpSGD\_$\ell2$) & Classifier (tpSGD\_$\ell2$) & Classifier (BP) \\ \hline
        Seq. MNIST (rows) & 10.32\% (10\%) & 19.02\% & 83.16\% & 79.96\% & 92.59\% \\ \hline
        Seq. MNIST (pixels) & 10.32\% (10\%) & 13.66\% & 68.60\% & 70.01\% & 82.04\% \\ \hline
        EEG Seizure & 51.57\% (50\%) & 80.87\% & 80.78\% & 80.87\% & 93.75\%  \\ \hline
        UCF101 & 1.22\% (0.99\%) & 52.64\% & 59.80\% & 59.75\% & 63.39\% \\ \hline
        Twitter (one-hot)  & 50.08\% (50\%) & 52.55\% & 67.04\% & 67.35\% & 69.94\% \\\hline
        Twitter (Word2Vec) & 50.08\% (50\%) & 64.35\% & 70.24\% & 68.48\% & 72.84\% \\ \hline
    \end{tabular}
    }
\caption{Results of training a two-layer stacked RNN with two-layer perceptron classifier. Numbers in parentheses are true random chance. We show the algorithm used to train each layer in parentheses where BP is traditional backpropagation.}
\label{tab:rnn_stacked}
\end{table*}

\begin{table*}[!htp]
\resizebox{1.0\textwidth}{!}{
    \centering
    \begin{tabular}{|c|c|c|c|c|c|}
    \hline
    \multirow{2}{*}{}
        Dataset & Random & Random RNN + & Trained RNN (DRTP) & Trained RNN (tpSGD\_$\ell1$) + & Trained RNN (BP) + \\
          & & Trained Classifier (tpSGD\_$\ell2$) & Classifier (tpSGD\_$\ell2$) & Classifier (tpSGD\_$\ell2$) & Classifier (BP) \\ \hline
        Seq. MNIST (rows) & 11.35\% (10\%) & 10.70\% & 83.13\% & 79.05\% & 92.10\% \\ \hline
        Seq. MNIST (pixels) & 11.35\% (10\%) & 10.32\% & 67.19\% & 68.01\% & 80.95\% \\ \hline
        EEG Seizure & 48.52\% (50\%) & 85.04\% & 84.09\% & 88.61\% & 90.53\% \\ \hline
        UCF101 & 1.01\% (0.99\%) & 56.78\% & 61.37\% & 61.63\% & 64.25\% \\ \hline
        Twitter (one-hot)  & 50.08\% (50\%) & 51.07\% & 65.00\% & 66.41\% & 71.65\% \\ \hline
    \multirow{2}{*}{}
        Twitter (Word2Vec) & 50.08\% (50\%) & 65.76\% & 70.31\% & 68.96\% & 69.74\% \\ \hline
    \end{tabular}
    }
\caption{Results of training a single-layer Bidirectional RNN with linear classifier. Numbers in parentheses are true random chance. We show the algorithm used to train each layer in parentheses where BP is traditional backpropagation.} 
\label{tab:rnn_bidirectional}
\end{table*}

\end{document}